\documentclass[runningheads]{llncs}
\usepackage[T1]{fontenc}
\usepackage{bm}
\usepackage{bbm}
\usepackage{graphicx,verbatim}
\usepackage{amssymb}
\usepackage{amsmath}
\usepackage{algorithm}
\usepackage{algpseudocode}
\usepackage[hidelinks]{hyperref}
\usepackage[table,xcdraw]{xcolor}
\usepackage{colortbl}
\usepackage{booktabs}
\usepackage{multirow}
\usepackage{makecell}
\usepackage{pifont}   
\newif\ifcameraready
\camerareadyfalse

\definecolor{gaincolor}{RGB}{0, 150, 80} 
\definecolor{losscolor}{RGB}{200, 0, 0}    
\definecolor{lightgrayline}{RGB}{220, 220, 220}
\definecolor{groupgray}{gray}{0.95}

\newcommand{\paperrowcolor}[1]{\rowcolor{#1}}
\newcommand{\papercellcolor}[1]{\cellcolor{#1}}
\newcommand{\paperhighlightrow}{\ifcameraready\rowcolor{gray!12}\else\rowcolor{green!5}\fi}
\newcommand{\papertextcolor}[2]{\ifcameraready #2\else\textcolor{#1}{#2}\fi}

\newcommand{\Ponly}{$\mathcal{P}$}
\newcommand{\PText}{$\mathcal{P}\!+\!\mathcal{T}$}

\newcommand{\stackgain}[2]{%
  \begin{tabular}{@{}c@{}} #1 \\[-2pt] \tiny{\papertextcolor{gaincolor}{$\uparrow$#2\%}} \end{tabular}%
}
\newcommand{\stackgainASD}[2]{%
  \begin{tabular}{@{}c@{}} #1 \\[-2pt] \tiny{\papertextcolor{gaincolor}{$\downarrow$#2\%}} \end{tabular}%
}

\def\eg{{\em e.g.}}
\begin{document}
\title{EchoPilot: Training-Free Ultrasound Video Segmentation via Scale-Space Semantic Prompting and Reliability-Gated Memory}
\titlerunning{EchoPilot}
%
\author{Ruiqiang Xiao\inst{1} \and
Zhaohu Xing\inst{1} \and
Yijun Yang\inst{1} \and
Zhenyan Han\inst{2} \and
Weiming Wang\inst{3} \and
Kaishun Wu\inst{1} \and
Lei Zhu\inst{1}\textsuperscript{\ding{41}}}
%
\authorrunning{Xiao et al.}
%
\institute{The Hong Kong University of Science and Technology (Guangzhou)\\
\email{leizhu@hkust-gz.edu.cn}
\and
Third Affiliated Hospital of Sun Yat-Sen University
\and
Hong Kong Metropolitan University}
\maketitle              
%


\begin{abstract}
Ultrasound video segmentation is clinically valuable yet difficult due to speckle noise, weak boundaries, and rapid anatomical deformation.
Recent promptable foundation models enable point-guided segmentation, but their direct deployment in ultrasound remains unreliable: a single point provides insufficient spatial context to resolve scale ambiguity, and greedy memory updates amplify early errors into severe temporal drift.
We present \textbf{EchoPilot}, a \textit{training-free} framework for ultrasound video segmentation under \textit{sparse first-frame interaction}, requiring only a single point click and an anatomical category name.
EchoPilot orchestrates a frozen medical vision--language model (VLM) for semantic localization, a vision foundation model (VFM) for dense geometric feature extraction, and a promptable video segmentor for mask prediction and propagation.
To resolve initialization ambiguity, we propose \textbf{Scale-Space Semantic Prompting}, which first selects an optimal contextual view via a parameter-free \textbf{S.E.E.D.}\ (Semantic Energy--Entropy Density) criterion, and then synthesizes geometrically precise auxiliary point prompts from dense foundation features without additional user interaction.
To reduce propagation drift, a \textbf{Reliability-Gated Memory} update is further introduced to selectively freeze the segmentor's memory bank under uncertain predictions, preventing error accumulation.
We also contribute the first \textbf{dynamic fetal placenta} ultrasound video segmentation dataset with 671 annotated frames.
Across three ultrasound video datasets, EchoPilot achieves state-of-the-art performance under the sparse-interactive setting, consistently outperforming training-free baselines and finetuned specialists.

\noindent\textbf{Project page:} \href{https://keeplearning-again.github.io/EchoPilot/}{https://keeplearning-again.github.io/EchoPilot/}.

\keywords{Video Object Segmentation \and Sparse Interaction \and Training-Free \and Ultrasound.}

\end{abstract}
\section{Introduction}
\begin{figure}[t] 
    \centering 
    \includegraphics[width=\textwidth]{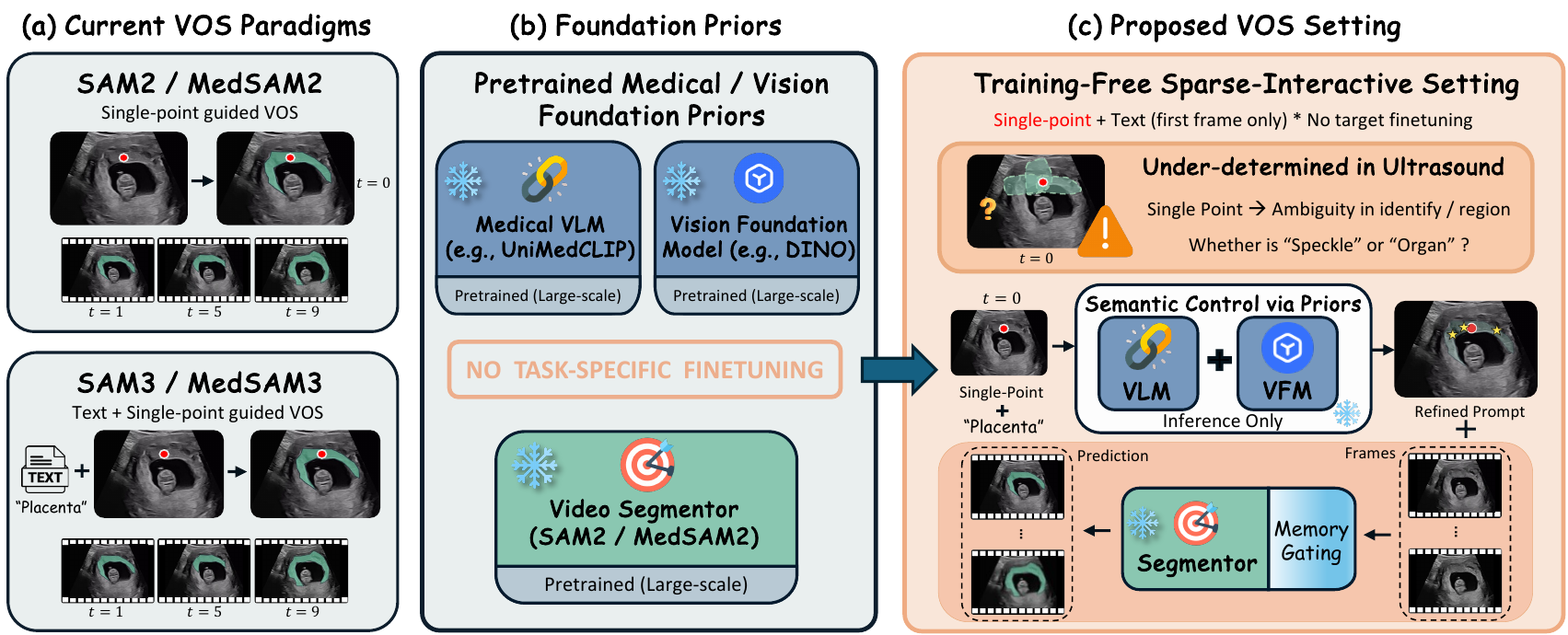} 
    \caption{\textbf{Setting and concept of EchoPilot.}
    (a)~Existing point-guided and text-guided VOS paradigms suffer from initialization ambiguity or require task-specific fine-tuning.
    (b)~EchoPilot orchestrates three frozen foundation priors---a medical VLM, a VFM, and a video segmentor---without any fine-tuning.
    (c)~Our category-anchored sparse interaction requires only a single point click and an anatomical category name on the first frame.} 
    \label{fig:teaser} 
\end{figure}

Video Object Segmentation (VOS) typically assumes a semi-supervised setting where a first-frame mask is propagated through the video~\cite{wang2018semi,gao2023deep,ding2023mose,cheng2024putting,perazzi2017learning,oh2019video}. 
Recent foundation models for interactive segmentation---notably SAM~2~\cite{sam2} and medical adaptations~\cite{wu2025medical,yan2025samed,medsam2,chen2025accelerating,masam2,kim2025echofm}---have reshaped this paradigm by enabling \emph{point-guided} video segmentation, 
where sparse clicks or boxes can replace dense masks. 
Text-conditioned variants like SAM~3~\cite{carion2025sam} and MedSAM~3~\cite{liu2025medsam3} further allow users to specify semantic intent through natural language, 
suggesting a practical path toward clinician-friendly VOS with minimal interaction (Fig.~\ref{fig:teaser}(a)).

Ultrasound is uniquely suited to real-time bedside assessment due to its portability, safety, and low cost~\cite{noble2006ultrasound}, yet it remains one of the most challenging modalities for automated segmentation, as images are inherently degraded by speckle noise, acoustic artifacts, weak and fluctuating boundaries~\cite{wagner2005statistics,camus}. 
Consequently, sparse point prompts on a single frame often \emph{under-determine} the segmentation task (Fig.~\ref{fig:teaser}(c), top): a click specifies a location but may still match multiple plausible structures. 
Although an anatomical category provides semantic intent, it does not specify the target size, surrounding context, or boundary shape in noisy ultrasound videos.
The appropriate contextual scale around the click---neither too large, which includes confusing nearby anatomy, nor too small, which removes useful cues for identification---therefore cannot be inferred from the point alone.
In the video setting, such initialization ambiguity is particularly detrimental: once a segmentor commits to an incorrect interpretation, greedy memory-based propagation reinforces and compounds the error over time, resulting in persistent object drift~\cite{kalal2011tracking,salti2012adaptive,sam2long}.

A common solution in ultrasound segmentation research is to collect task-specific video annotations and fine-tune models for each anatomy, protocol, or device domain. 
However, annotation is expensive, susceptible to inter-observer variability~\cite{willemink2020preparing}, and fragile to domain shift across hardware and imaging settings~\cite{liu2020deep,kelly2019key}. 
While training-free VOS has been explored under dense first-frame masks in surgical instrument segmentation~\cite{masam2}, the combination of \emph{training-free inference} with \emph{sparse first-frame interaction} remains largely unaddressed for ultrasound video segmentation.
We therefore study \textbf{training-free, sparse-interactive ultrasound VOS}, where all models are kept frozen at inference time and tracking is driven only by a single first-frame point click and an anatomical category name.

We propose EchoPilot, a training-free pipeline that targets the two dominant failure modes in this setting: \emph{initialization ambiguity} and \emph{error amplification during propagation}.
The key observation is that different pretrained priors capture complementary cues for sparse-interactive VOS (Fig.~\ref{fig:teaser}(b)):
a medical vision--language model (VLM) provides text-grounded semantic evidence about \emph{what} to segment, a vision foundation model (VFM) supplies dense appearance features to refine \emph{where} to prompt,
and a promptable video segmentor handles temporal correspondence for mask propagation.
A Reliability-Gated Memory update policy further regulates the memory bank, preventing uncertain predictions from contaminating subsequent frames and reducing drift.
EchoPilot is directly compatible with existing point-guided video segmentors (\eg, SAM~2~\cite{sam2}, MedSAM~2~\cite{medsam2}).
We also curate the first dynamic fetal placenta ultrasound VOS dataset with 671 annotated frames. We evaluate on three ultrasound datasets spanning diverse anatomies, boundary conditions, and motion dynamics. 

Our contributions are as follows: 
(1) we introduce \textbf{EchoPilot}, a plug-and-play framework for training-free ultrasound VOS from a single first-frame click and an anatomical category name; 
(2) we propose \textbf{Scale-Space Semantic Prompting} with \textbf{S.E.E.D.}, a parameter-free energy--entropy criterion for VLM-guided scale selection, together with \textbf{Reliability-Gated Memory} to reduce propagation drift; 
(3) we curate the first dynamic fetal placenta ultrasound VOS dataset and demonstrate consistent gains over training-free baselines and finetuned specialists across three ultrasound video datasets.

\section{Method}
\subsection{Problem Setup and Overview}
Given an ultrasound video $\{I_t\}_{t=0}^{T-1}$, a user provides a single positive point $p_0$ and an anatomical category name $\mathcal{C}$ on the first frame $I_0$. Our goal is to predict a segmentation mask $M_t$ for each frame $t$ in a \textbf{training-free} manner.
The category name is a task-level descriptor (\eg, ``placenta'') fixed per clinical protocol. After initialization, the segmentor operates solely on \textbf{point prompts}.

\begin{figure}[t] 
    \centering 
    \includegraphics[width=1.\textwidth]{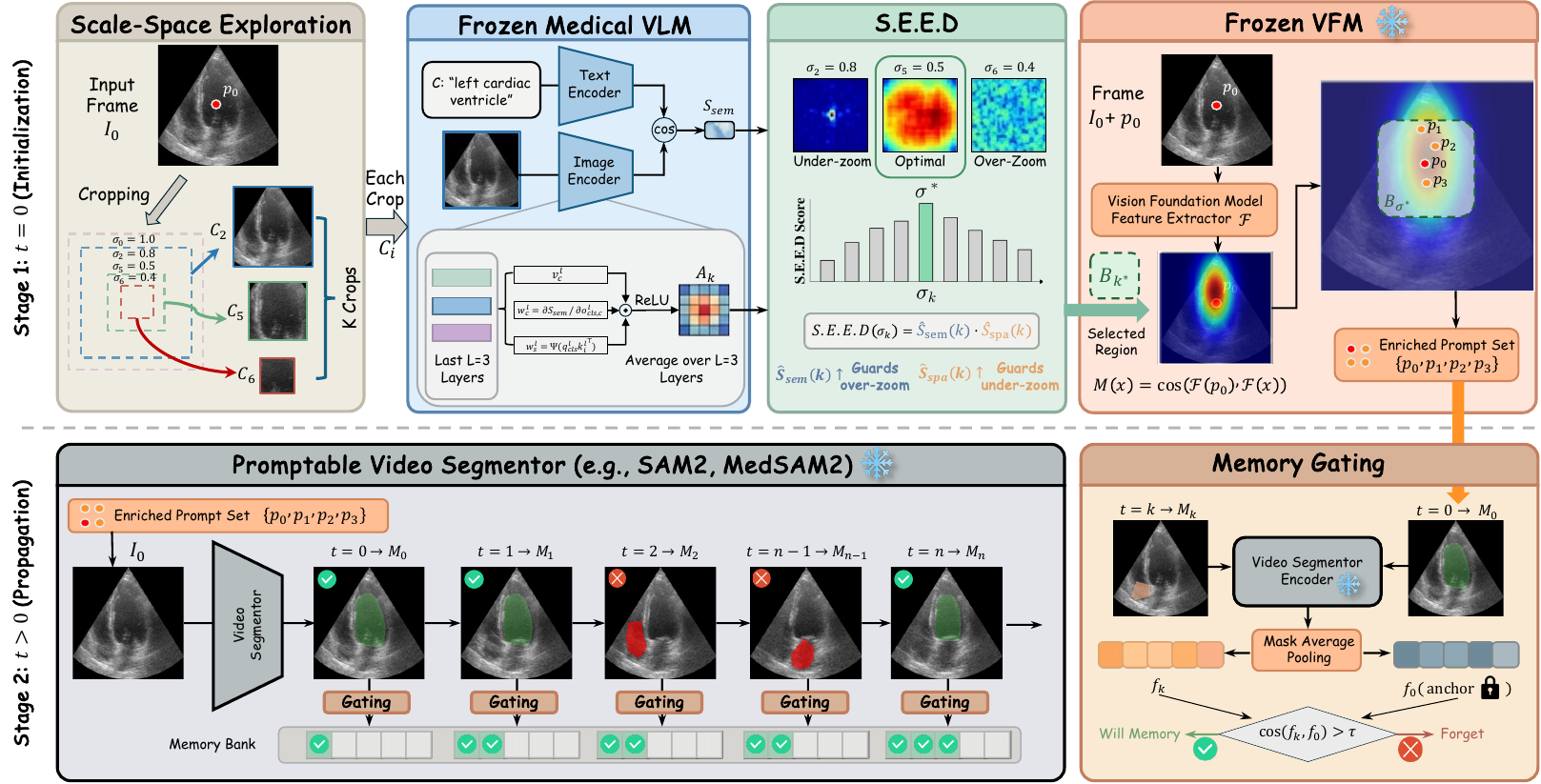} 
    \caption{\textbf{Overview of EchoPilot.}
    At $t{=}0$ (Stage~I), the S.E.E.D.\ criterion selects an optimal contextual scale via a frozen medical VLM, and a VFM refines point prompts within the selected region.
    For $t{>}0$ (Stage~II), a reliability-gated write policy controls which predicted frames enter the SAM~2/MedSAM~2 memory bank, suppressing error accumulation.}
    \label{fig:method} 
\end{figure}

As illustrated in Fig.~\ref{fig:method}, EchoPilot consists of two stages.
\textbf{Stage~I} ($t{=}0$) resolves initialization ambiguity by determining \emph{what} to segment and \emph{where} to prompt: it identifies a contextual scale where the target anatomy is semantically recognizable and spatially dominant, then synthesizes geometrically precise auxiliary point prompts within the selected region.
\textbf{Stage~II} ($t{>}0$) stabilizes temporal propagation by controlling \emph{when} to update the segmentor's memory bank, preventing unreliable predictions from contaminating subsequent frames.

\subsection{Stage~I: Scale-Space Semantic Prompting}
A single point click is ambiguous in ultrasound since the visible context depends on the observation scale: too-narrow views are dominated by speckle, while too-wide views introduce competing structures.
We search for a scale where the target anatomy is both semantically identifiable and spatially prominent.

\subsubsection{Multi-scale Crops and VLM-based Attribution.}
We generate $K$ center crops $\{C_k\}_{k=1}^{K}$ around $p_0$ at decreasing scale factors $\sigma_k$.
For each crop, a frozen medical VLM encodes the image and category name via its image encoder $\mathcal{E}_{img}$ and text encoder $\mathcal{E}_{txt}$, yielding a clamped cosine similarity:
\begin{equation}\label{eq:sim}
  \mathcal{S}_{sem}(k)
  = \max\!\bigl(0,\;
    \cos\bigl(\mathcal{E}_{img}(C_k),\;
              \mathcal{E}_{txt}(\mathcal{C})\bigr)\bigr).
\end{equation}
This score indicates \emph{whether} the crop matches the target but not \emph{where}.
Following~\cite{zhao2024gradient}, we extract a text-conditioned spatial attribution map $A_k\!\in\!\mathbb{R}^{h\times w}$ (where $h{\times}w$ is the VLM spatial token grid)
by backpropagating $\mathcal{S}_{sem}(k)$ through the VLM's attention layers.
For spatial token $i$ at layer $l$:
\begin{equation}\label{eq:heatmap}
  A_k^{(l)}(i)=\mathrm{ReLU}\!\left(
      \sum_c
      \underbrace{\left(\frac{\partial \mathcal{S}_{sem}}{\partial o_{\mathrm{cls},c}^{(l)}}\right)}_
      {w_c^{(l)}\;\text{(channel)}}
      \;\cdot\;
      \underbrace{\Psi\!\left(q_{\mathrm{cls}}^{(l)}\,
      {k_i^{(l)}}^\top\right)}_
      {w_s^{(l)}\;\text{(spatial)}}
      \;\cdot\; v_{i,c}^{(l)}
  \right),
\end{equation}
where $w_c^{(l)}$ is the channel importance weight (gradient of $\mathcal{S}_{sem}$ with respect to the class-token attention output $o_{\mathrm{cls},c}^{(l)}$), $w_s^{(l)}(i)$ is the spatial importance weight (normalized attention affinity between the class-token query $q_{\mathrm{cls}}^{(l)}$ and the key $k_i^{(l)}$ of token $i$), and $v_{i,c}^{(l)}$ is the corresponding value at channel $c$.
$\Psi(\cdot)$ denotes affinity normalization that suppresses sparse responses~\cite{zhao2024gradient}.
We aggregate over the last $L$ layers ($A_k = \frac{1}{L}\sum_{l} A_k^{(l)}$) to reduce layer-specific localization noise.

\subsubsection{S.E.E.D.\ Contextual Scale Selection.}
$\mathcal{S}_{sem}$ alone cannot distinguish under-zoomed crops where the anatomy is recognizable but occupies only a small fraction of the view.
We introduce \textbf{S.E.E.D.}\ (Semantic Energy--Entropy Density), a parameter-free criterion that couples average attribution strength with spatial dispersion under a semantic constraint.
Together, energy density suppresses diffuse low-confidence responses, while normalized entropy discourages overly collapsed attribution patterns that often arise from speckle or isolated edges.
We normalize $A_k$ into a distribution $P_k(i)=A_k(i)/(\sum_j A_k(j)+\epsilon)$ over $N_p=h{\times}w$ tokens, where $\epsilon$ is a small stability constant:
\begin{equation}
  \mathcal{S}_{spa}(k)=
  \underbrace{\left(\frac{1}{N_p}\sum_{i}A_k(i)\right)}_{\text{Energy Density}}
  \;\cdot\;
  \underbrace{\left(\frac{-\sum_{i}P_k(i)\log P_k(i)}{\log N_p}\right)}_{\text{Normalized Entropy}}.
\end{equation}
Over-zoomed crops tend to penalize $\mathcal{S}_{sem}$ or collapse attribution onto local artifacts, whereas under-zoomed crops reduce attribution density by spreading evidence over irrelevant anatomy.
After min--max normalization ($\hat{\mathcal{S}}$), the optimal scale is
\begin{equation}
  k^{*}=\arg\max_{k}\;
  \Big(\hat{\mathcal{S}}_{sem}(k)\cdot\hat{\mathcal{S}}_{spa}(k)\Big).
\end{equation}
The bounding box $B_{k^*}$ centered at $p_0$ at scale $\sigma_{k^*}$ serves as the observation window for prompt refinement.

\subsubsection{Prompt Refinement with a VFM.}
The VLM attribution operates on a coarse token grid, so we refine point selection using dense features from a vision foundation model.
Let $\mathcal{G}(\cdot)\in\mathbb{R}^{d}$ denote the VFM feature at a spatial location in $I_0$.
We compute a dense cosine similarity map anchored at the user click:
\begin{equation}\label{eq:vfm_sim}
  \mathcal{M}(x)=\cos\!\bigl(\mathcal{G}(p_0),\,\mathcal{G}(x)\bigr), \quad \forall x\in I_0.
\end{equation}
Within $B_{k^*}$, we extract up to $3$ local maxima via non-maximum suppression (radius $r$) as auxiliary positive prompts $\{p_1,p_2,p_3\}$.
This cap preserves the sparse-interaction setting while still covering elongated or non-convex anatomical structures. NMS avoids redundant peaks, and the S.E.E.D.-selected window suppresses weak similarity maxima from background tissue.
Restricting refinement to this window suppresses background distractors, so explicit negative prompts are unnecessary.
The final prompt set $\{p_0,p_1,p_2,p_3\}$ is passed to the video segmentor.

\subsection{Stage~II: Reliability-Gated Memory Update}

Promptable video segmentors such as SAM~2 maintain a fixed-size
memory bank and, by default, update it greedily by writing each predicted frame in temporal order.
Recent work has identified this strategy as a key source of error
accumulation in long or ambiguous videos~\cite{sam2long,masam2}.
In ultrasound, severe non-rigid deformation, acoustic shadowing, and transient artifacts often yield unreliable masks;
once stored, these corrupt subsequent predictions and cause compounding drift.

We therefore introduce a training-free, conservative feature-consistency write gate that decides, per frame,
whether the prediction provides sufficiently stable evidence to enter the memory bank.
Specifically, the gate reuses the memory-encoder feature map
$\mathcal{F}_t^{\mathrm{mem}}\!\in\!\mathbb{R}^{h\times w\times d}$
already computed by SAM~2 at every frame, introducing \emph{zero} additional forward passes.
Given the predicted mask probabilities
$M_t \in [0,1]^{h\times w}$, we extract a target-specific
descriptor $\mathbf{f}_t \in \mathbb{R}^d$ via masked average
pooling:
\begin{equation}\label{eq:map}
  \mathbf{f}_t
  = \frac{\sum_{x} M_t(x)\,\mathcal{F}_t^{\mathrm{mem}}(x)}
         {\sum_{x} M_t(x) + \epsilon}.
\end{equation}
We then measure feature consistency with a first-frame anchor $\mathbf{f}_0$, computed once from the Stage~I initialization:
\begin{equation}\label{eq:gate}
  g_t = \cos(\mathbf{f}_t,\,\mathbf{f}_0),
  \qquad W_t = \mathbb{I}[g_t > \tau],
\end{equation}
where $\mathbb{I}[\cdot]$ is the indicator function and $\tau$ is a reliability threshold.
If $W_t{=}1$, the frame is written into the memory bank; otherwise the
write is skipped to prevent unreliable evidence from contaminating
the bank.
Since legitimate appearance changes from deformation, probe motion, or acoustic shadowing can lower consistency, the gate is one-sided: high consistency permits an update, while low consistency triggers a skip rather than forcing a correction---favoring stable propagation over aggressive appearance adaptation.
This mechanism preserves SAM~2's architecture and weights unchanged,
yielding a plug-and-play drop-in replacement for the default greedy memory update.

\section{Experiments and Results}
\begin{table}[t]
    \centering
    \caption{\textbf{Quantitative comparison on three ultrasound VOS datasets.} Methods are grouped by pretrained backbone weights (SAM2 and MedSAM2). $\mathcal{P}$ and $\mathcal{T}$ denote point and text prompts, respectively. \textbf{Bold} and \underline{underline} mark the best/second-best results within each weight group. Percentages below EchoPilot entries indicate relative improvement over the corresponding base segmentor (SAM2 or MedSAM2).}
    \label{tab:main_results}
    \setlength{\tabcolsep}{2.2pt}
    \resizebox{\textwidth}{!}{%
    \begin{tabular}{l l c ccc ccc ccc}
    \toprule
    \multirow{2}{*}{\textbf{Method}} & \multirow{2}{*}{\textbf{Venue}} & \multirow{2}{*}{\textbf{Prompt}} & \multicolumn{3}{c}{\textbf{CAMUS}} & \multicolumn{3}{c}{\textbf{Breast Lesion}} & \multicolumn{3}{c}{\textbf{Placenta}} \\
    \cmidrule(lr){4-6} \cmidrule(lr){7-9} \cmidrule(lr){10-12}
     & & & $\mathcal{D} \uparrow (\%)$ & $\mathcal{A} \downarrow$ & $\mathcal{F} \uparrow (\%)$ & $\mathcal{D} \uparrow (\%)$ & $\mathcal{A} \downarrow$ & $\mathcal{F} \uparrow (\%)$ & $\mathcal{D} \uparrow (\%)$ & $\mathcal{A} \downarrow$ & $\mathcal{F} \uparrow (\%)$ \\
    \midrule
    \paperrowcolor{gray!5}
    MedSAM~3 & arXiv'25 & \PText & 67.15 & 13.98 & 15.24 & 56.93 & 43.31 & 16.58 & 20.69 & 140.16 & 5.39 \\
    \midrule
    \multicolumn{12}{l}{\papercellcolor{groupgray}\textit{\textbf{Pretrained Weights: SAM2}}} \\
    SAM2 & ICLR'25 & \Ponly & 28.41 & 56.84 & 0.80 & \underline{56.00} & 56.54 & \underline{19.83} & 16.74 & \underline{267.24} & 0.89 \\
    MA-SAM2 & MICCAI'25 & \Ponly & 28.43 & 56.81 & \underline{0.81} & 55.76 & 56.61 & 19.67 & 16.75 & 267.26 & \underline{0.90} \\
    SAM2Long & ICCV'25 & \Ponly & \underline{28.44} & \underline{56.65} & 0.80 & 55.41 & \underline{54.96} & 19.56 & \underline{16.76} & 268.99 & 0.89 \\
    \paperhighlightrow
    \textbf{EchoPilot} & - & \PText & 
    \stackgain{\textbf{34.09}}{19.99} & \stackgainASD{\textbf{28.86}}{49.23} & \stackgain{\textbf{3.21}}{301.25} & 
    \stackgain{\textbf{63.38}}{13.18} & \stackgainASD{\textbf{22.98}}{59.36} & \stackgain{\textbf{21.35}}{7.67} & 
    \stackgain{\textbf{39.74}}{137.40} & \stackgainASD{\textbf{84.96}}{68.21} & \stackgain{\textbf{5.92}}{565.17} \\
    \midrule
    \multicolumn{12}{l}{\papercellcolor{groupgray}\textit{\textbf{Pretrained Weights: MedSAM2}}} \\
    MedSAM2 & arXiv'25 & \Ponly & 90.83 & 2.29 & 77.34 & 61.24 & 86.12 & 23.63 & 33.52 & 129.56 & 3.27 \\
    MA-SAM2 & MICCAI'25 & \Ponly & 90.85 & 2.28 & \underline{77.37} & 61.15 & 86.14 & 23.25 & 33.54 & 129.54 & \underline{3.29} \\
    SAM2Long & ICCV'25 & \Ponly & \underline{91.01} & \underline{2.23} & \textbf{78.91} & \underline{66.73} & \underline{47.00} & \textbf{26.10} & \underline{34.92} & \underline{120.38} & \textbf{3.48} \\
    \paperhighlightrow
    \textbf{EchoPilot} & - & \PText & 
    \stackgain{\textbf{95.31}}{4.93} & \stackgainASD{\textbf{0.83}}{63.76} & \stackgain{77.35}{0.01} & 
    \stackgain{\textbf{68.44}}{11.76} & \stackgainASD{\textbf{28.20}}{67.25} & \stackgain{\underline{23.91}}{1.18} & 
    \stackgain{\textbf{38.87}}{15.96} & \stackgainASD{\textbf{63.40}}{51.07} & \stackgain{3.28}{0.31} \\
    \bottomrule
    \end{tabular}%
    }
\end{table}

\noindent\textbf{Datasets and Evaluation Metrics.}
We evaluate on three ultrasound video segmentation datasets:
\textit{CAMUS}~\cite{camus},
\textit{Breast Lesion}~\cite{breast},
and \textit{Placenta}, a newly collected fetal placenta VOS dataset.
We report Dice coefficient ($\mathcal{D}$),
Average Surface Distance ($\mathcal{A}$),
and mean F-boundary score ($\mathcal{F}$)~\cite{perazzi2016benchmark} as evaluation metrics.

\noindent\textbf{Implementation Details.}
All experiments are conducted in a training-free setting. To simulate clinical point prompts, we randomly sample a single positive point from the ground-truth mask on the first frame of each video; we fix random seeds to $\{2025, 2026, 42\}$
and report results averaged across all seeds.
For VLM, we adopt BioMedCLIP (ViT-B)~\cite{biomedclip}, a biomedical vision--language foundation model pretrained on large-scale diverse biomedical image--text pairs.
For VFM, we use DINOv3 (ViT-Plus)~\cite{dinov3}.
The NMS radius for auxiliary prompt selection is $r{=}6$ pixels, and the memory-gate threshold in Stage~II is $\tau{=}0.5$ as default.

\noindent\textbf{Results and visualization.}
We compare EchoPilot against training-free prompting strategies (SAM2~\cite{sam2}, MA-SAM2~\cite{masam2}, SAM2Long~\cite{sam2long}) and finetuned medical segmentors (MedSAM2~\cite{medsam2}, MedSAM~3~\cite{liu2025medsam3}).
The prompt column reflects each method's native interface: SAM2-family baselines are point-guided, whereas EchoPilot and MedSAM~3 can consume the anatomical category name at initialization.
We therefore interpret Tab.~\ref{tab:main_results} as an end-to-end comparison under the target sparse-interactive setting, and use the ablations below to separate the effects of VLM-guided initialization, VFM-based auxiliary prompts, and memory gating under matched segmentor weights.

As shown in Tab.~\ref{tab:main_results}, EchoPilot achieves the best Dice and ASD across all three datasets under both weight families. The gains are most pronounced on the challenging Placenta dataset,
where EchoPilot improves Dice by $+$23.0\% absolute over the SAM2 baseline (16.74$\to$39.74) and reduces ASD by 68\%. With MedSAM2 weights, EchoPilot reaches 95.31\% Dice on CAMUS
with an ASD of only 0.83, approaching clinical-grade accuracy. Notably, despite being fully training-free, EchoPilot consistently outperforms the finetuned MedSAM~3 by a large margin on all
datasets, confirming that scale-space semantic prompting and reliability-gated memory update can effectively substitute for task-specific supervision.
Under MedSAM2 weights, the boundary $\mathcal{F}$-score of EchoPilot still surpasses the MedSAM2 baseline on all datasets, while SAM2Long achieves marginally higher $\mathcal{F}$ on some splits; this is expected because the reliability gate favors conservative memory writes that stabilize region overlap (Dice/ASD) at the cost of slightly smoother boundaries.

\begin{figure}[t]
    \centering 
    \includegraphics[width=\textwidth]{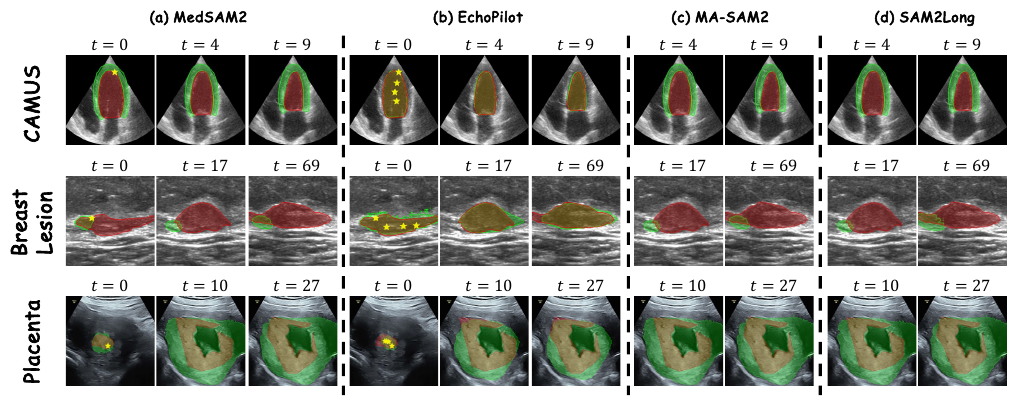} 
    \caption{\textbf{Qualitative comparison across time steps on three ultrasound VOS datasets.} Each row shows segmentation masks at selected frames for (a)~MedSAM2, (b)~EchoPilot (Ours), (c)~MA-SAM2, and (d)~SAM2Long. \papertextcolor{red}{Red} contours indicate ground-truth boundaries and \papertextcolor{green}{green} contours indicate predictions. Yellow stars in~(b) mark the auxiliary point prompts synthesized by Stage~I. Baseline methods exhibit progressive mask drift and boundary leakage over time, whereas EchoPilot maintains stable and accurate delineation throughout the sequence.}
    \label{fig:qualitative} 
\end{figure}

Fig.~\ref{fig:qualitative} reveals two consistent advantages of EchoPilot:
(i)~the auxiliary prompts from Stage~I yield well-aligned masks from the very first frame, and
(ii)~on longer sequences (Breast Lesion, Placenta), the predicted contours remain anchored to the target with markedly less boundary leakage and shape drift than all baselines.

\begin{table}[!htbp]
    \centering
    \caption{\textbf{Ablation of Stage~I initialization on all three datasets.} We compare the base segmentor (single random click), VFM-only prompt selection (without VLM scale guidance), and the full EchoPilot pipeline instantiated with two different medical VLM backbones. $\Delta$ denotes the change relative to the base.}
    \label{tab:ablation_stage1}
    \setlength{\tabcolsep}{2.2pt}
    \resizebox{\textwidth}{!}{%
    \begin{tabular}{l cccc cccc cccc}
    \toprule
    \multirow{2}{*}{\textbf{Stage-I Variant}} &
    \multicolumn{4}{c}{\textbf{CAMUS}} &
    \multicolumn{4}{c}{\textbf{Breast Lesion}} &
    \multicolumn{4}{c}{\textbf{Fetal Placenta}} \\
    \cmidrule(lr){2-5}\cmidrule(lr){6-9}\cmidrule(lr){10-13}
    & $\mathcal{D}\uparrow(\%)$ & $\mathcal{A}\downarrow$ & $\Delta\mathcal{D}$ (\%) & $\Delta\mathcal{A}$ 
    & $\mathcal{D}\uparrow(\%)$ & $\mathcal{A}\downarrow$ & $\Delta\mathcal{D}$ (\%) & $\Delta\mathcal{A}$ 
    & $\mathcal{D}\uparrow(\%)$ & $\mathcal{A}\downarrow$ & $\Delta\mathcal{D}$ (\%) & $\Delta\mathcal{A}$ \\
    \midrule
    \multicolumn{13}{l}{\papercellcolor{groupgray}\textit{\textbf{Pretrained Weights: SAM2}}} \\
    SAM2 (Base; 1-click random) 
    & 28.41 & 56.84 & -- & -- 
    & 56.00 & 56.54 & -- & -- 
    & 16.74 & 267.24 & -- & -- \\
    VFM-only 
    & 30.52 & 30.82 & \papertextcolor{gaincolor}{+2.11} & \papertextcolor{gaincolor}{-26.02}
    & 54.87 & 29.38 & \papertextcolor{losscolor}{-1.13} & \papertextcolor{gaincolor}{-27.16}
    & 39.73 & 87.34 & \papertextcolor{gaincolor}{+22.99} & \papertextcolor{gaincolor}{-179.90} \\
    \midrule
    UniMedCLIP+VFM (EchoPilot)
    & 29.98 & 31.16 & \papertextcolor{gaincolor}{+1.57} & \papertextcolor{gaincolor}{-25.68}
    & 57.78 & 27.61 & \papertextcolor{gaincolor}{+1.78} & \papertextcolor{gaincolor}{-28.93}
    & 39.04 & 88.33 & \papertextcolor{gaincolor}{+22.30} & \papertextcolor{gaincolor}{-178.91} \\
    BioMedCLIP+VFM (EchoPilot) 
    & \textbf{34.09} & \textbf{28.86} & \papertextcolor{gaincolor}{+5.68} & \papertextcolor{gaincolor}{-27.98}
    & \textbf{63.38} & \textbf{22.98} & \papertextcolor{gaincolor}{+7.38} & \papertextcolor{gaincolor}{-33.56}
    & \textbf{39.74} & \textbf{84.96} & \papertextcolor{gaincolor}{+23.00} & \papertextcolor{gaincolor}{-182.28} \\
    \midrule
    \multicolumn{13}{l}{\papercellcolor{groupgray}\textit{\textbf{Pretrained Weights: MedSAM2}}} \\
    MedSAM2 (Base; 1-click random) 
    & 90.83 & 2.29 & -- & -- 
    & 61.24 & 86.12 & -- & -- 
    & 33.52 & 129.56 & -- & -- \\
    VFM-only 
    & 78.32 & 2.65 & \papertextcolor{losscolor}{-12.51} & \papertextcolor{losscolor}{+0.36}
    & 67.72 & 24.92 & \papertextcolor{gaincolor}{+6.48} & \papertextcolor{gaincolor}{-61.20}
    & 37.81 & 64.50 & \papertextcolor{gaincolor}{+4.29} & \papertextcolor{gaincolor}{-65.06} \\
    UniMedCLIP+VFM (EchoPilot)
    & 84.69 & 2.08 & \papertextcolor{losscolor}{-6.14} & \papertextcolor{gaincolor}{-0.21}
    & 68.64 & 21.22 & \papertextcolor{gaincolor}{+7.40} & \papertextcolor{gaincolor}{-64.90}
    & 38.54 & 63.53 & \papertextcolor{gaincolor}{+5.02} & \papertextcolor{gaincolor}{-66.03} \\
    BioMedCLIP+VFM (EchoPilot) 
    & \textbf{95.31} & \textbf{0.83} & \papertextcolor{gaincolor}{+4.48} & \papertextcolor{gaincolor}{-1.46}
    & \textbf{68.44} & 28.20 & \papertextcolor{gaincolor}{+7.20} & \papertextcolor{gaincolor}{-57.92}
    & \textbf{38.87} & \textbf{63.40} & \papertextcolor{gaincolor}{+5.35} & \papertextcolor{gaincolor}{-66.16} \\
    \bottomrule
    \end{tabular}%
    }
\end{table}

\begin{figure}[t] 
    \centering 
    \includegraphics[width=\textwidth]{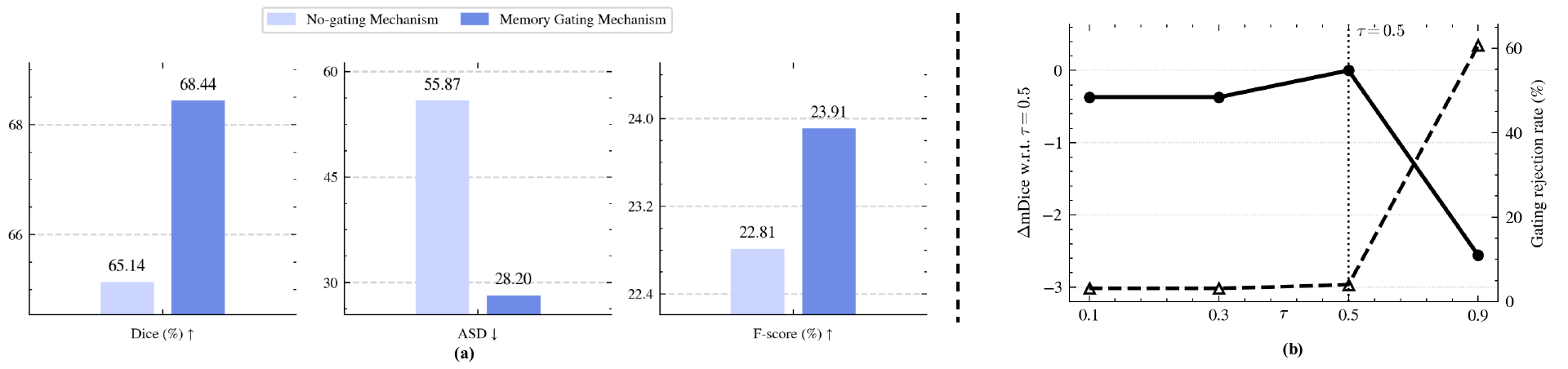} 
    \caption{\textbf{Ablation of the Reliability-Gated Memory Update on the Breast Lesion dataset (MedSAM2 weights).}
    \textbf{(a)}~Enabling the gating mechanism substantially reduces ASD from 55.87 to 28.20 by preventing artifact-corrupted predictions from entering the memory bank.
    \textbf{(b)}~Sensitivity analysis of the threshold $\tau$: performance remains stable across $\tau \in [0.1, 0.5]$, while an overly aggressive threshold ($\tau{=}0.9$) causes a rejection rate exceeding 60\%, discarding valuable temporal context and degrading Dice.}
    \label{fig:ablation} 
\end{figure}

\noindent\textbf{Ablation Studies.}
Table~\ref{tab:ablation_stage1} shows that VFM-only prompting is unstable ($-$12.5\% Dice on CAMUS, MedSAM2), whereas adding a frozen VLM as a semantic anchor yields consistent gains with both UniMedCLIP~\cite{unimedclip} and BioMedCLIP~\cite{biomedclip}, confirming that the improvement stems from the scale-space prompting formulation rather than a specific VLM.
Fig.~\ref{fig:ablation} validates the reliability gate: it halves ASD on Breast Lesion (55.87$\to$28.20) and is robust for $\tau \in [0.1, 0.5]$, while $\tau{=}0.9$ over-rejects frames and degrades Dice.

\section{Conclusion}
We presented \textbf{EchoPilot}, a training-free framework for ultrasound video object segmentation that requires only a single point click and an anatomical category name.
By coupling Scale-Space Semantic Prompting with Reliability-Gated Memory update, EchoPilot resolves initialization ambiguity and suppresses propagation drift without any task-specific fine-tuning.
Experiments on CAMUS, Breast Lesion, and a newly curated fetal placenta dataset show that EchoPilot consistently outperforms both training-free baselines and finetuned specialists under two video segmentor models, while the VLM and VFM priors are queried only once on the first frame, preserving practical throughput.

\begin{credits}
\subsubsection{\ackname}
This work was supported by the Guangdong Provincial Key Laboratory of Integrated Communication, Sensing and Computation for Ubiquitous Internet of Things (No. 2023B1212010007), Guangdong Science and Technology Department (2024ZDZX2004), the National Natural Science Foundation of China (NSFC) Grant (No. 62472366), a grant from the Research Grants Council of the Hong Kong Special Administrative Region, China (No. UGC/FDS16/E02/23), and AI Research and Learning Base of Urban Culture under Project 2023WZJD008.
\end{credits}

\bibliographystyle{splncs04}
\bibliography{reference.bib}


\end{document}